\documentclass[twoside]{article}

\usepackage[accepted]{aistats2017}
\usepackage[dvipsnames]{xcolor}
\usepackage{natbib}
\usepackage{url}
\usepackage{amsmath}
\usepackage{amsfonts}
\usepackage{stackengine}
\usepackage{graphicx}
\usepackage{subcaption}

\usepackage{hyperref}
\hypersetup{
  colorlinks   = true,
  citecolor    = blue,
  linkcolor     = blue,
}
\usepackage{cleveref}

\renewcommand{\deg}{\ensuremath{^\circ}}

\newcommand{\N}{\mathcal{N}}
\renewcommand{\v}[1]{\mathbf{#1}}
\newcommand{\ve}{\v{\varepsilon}}

\begin{document}

%

%

\twocolumn[

\aistatstitle{Signal-based Bayesian Seismic Monitoring}

\aistatsauthor{ David A. Moore \And Stuart J. Russell}
\aistatsaddress{University of California, Berkeley\\{\tt
    dmoore@cs.berkeley.edu} \And University of California, Berkeley\\{\tt russell@cs.berkeley.edu}}]

\begin{abstract}
  Detecting weak seismic events from noisy sensors is a difficult
  perceptual task. We formulate this task as Bayesian inference and
  propose a generative model of seismic events and signals across a
  network of spatially distributed stations. Our system, SIGVISA, is
  the first to directly model seismic waveforms, allowing it to
  incorporate a rich representation of the physics underlying the
  signal generation process. We use Gaussian processes over wavelet parameters to predict
  detailed waveform fluctuations based on historical events, while
  degrading smoothly to simple parametric envelopes in regions with no
  historical seismicity. Evaluating on data from the western US, we
  recover three times as many events as previous work, and reduce mean
  location errors by a factor of four while greatly increasing 
  sensitivity to low-magnitude events.
\end{abstract}

\section{Introduction}

The world contains structure: objects with dynamics governed by
physical law. Intelligent systems must infer this structure
from noisy, jumbled, and lossy sensory data. Bayesian statistics
provides a natural framework for designing such systems: given a prior distribution
$p(z)$ on underlying descriptions of the world, and a forward model $p(x|z)$
describing the process by which observations are generated,
mathematical probability defines the posterior $p(z|x) \propto p(z)p(x|z)$ over worlds
given observed data. Bayesian generative models have recently shown
exciting results in applications ranging from visual scene
understanding \citep{kulkarni2015picture,eslami2016attend} to
inferring celestial bodies \citep{regier2015celeste}.

In this paper we apply the Bayesian framework directly to a
challenging perceptual task: monitoring seismic events from a network
of spatially distributed sensors. This task is motivated by the Comprehensive Test Ban
Treaty (CTBT), which bans testing of nuclear weapons and provides for
the establishment of an International Monitoring System (IMS) to
detect nuclear explosions, primarily from the seismic signals that
they generate. The inadequacy of existing monitoring systems was cited as a
factor in the US Senate's 1999 decision not to ratify the treaty. 

Our system, SIGVISA (Signal-based Vertically Integrated Seismic
Analysis), consists of a generative probability model of seismic
events and signals, with interpretable latent variables for 
physically meaningful quantities such as the arrival times and
amplitudes of seismic phases. A previous system, NETVISA \citep{arora2013net}, assumed that
signals had been preprocessed into discrete detections; we extend this
by directly modeling seismic waveforms. This allows our model to
capture rich physical structure such as path-dependent modulation as
well as predictable travel times and attenuations. Inference in our model
recovers a posterior over event histories directly from
waveform traces, combining top-down with bottom-up processing to
produce a joint interpretation of all observed data. In particular,
Bayesian inference provides a principled approach to combining
evidence from phase travel times and waveform correlations, a previously unsolved problem in
seismology. 



A full description of our system is given by \citet{moore2016signal};
this paper describes the core model and key points of training and
inference algorithms. Evaluating against existing systems for seismic
monitoring, we show that SIGVISA significantly increases event recall
at the same precision, detecting many additional low-magnitude events
while reducing mean location error by a factor of four. Initial
results indicate we also perform as well or better than existing
systems at detecting events with no nearby historical seismicity.

\begin{figure}
\centering
\includegraphics[width=.45\textwidth]{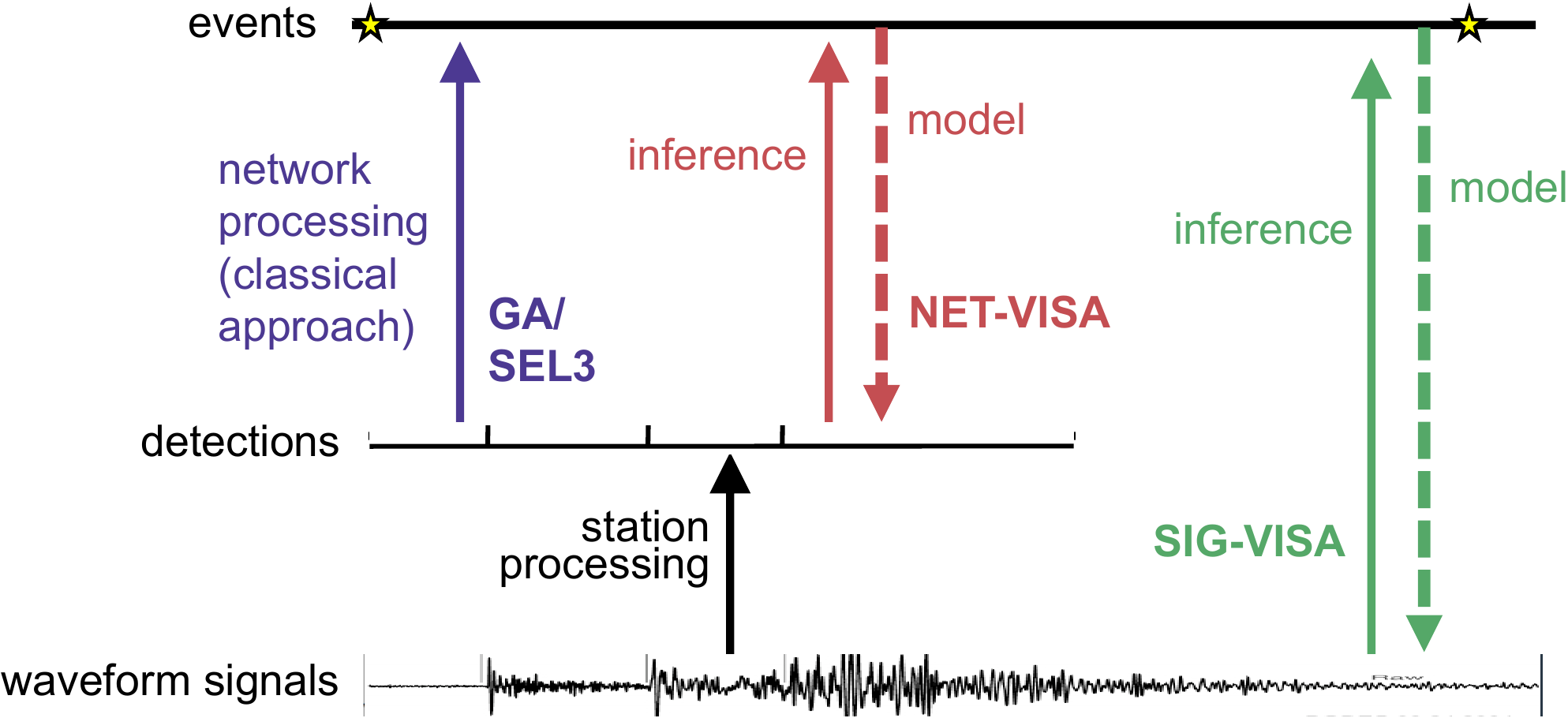}
\caption{High level structure of traditional detection-based monitoring (GA),
  Bayesian monitoring (NETVISA), and signal-based
  Bayesian monitoring (SIGVISA, this work). Compared to
  detection-based approaches, inference in a signal-based model
  incorporates rich information from seismic waveforms.}
\label{fig:monitoring_comparison}
\end{figure}

\section{Seismology background}


\begin{figure}
\centering
    \includegraphics[width=0.45\textwidth]{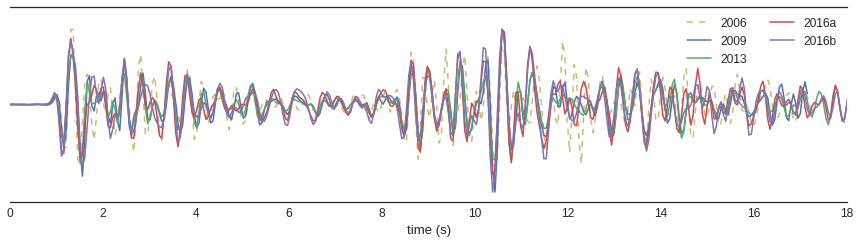}
    \caption{Aligned seismic waveforms (station MDJ,
      amplitude-normalized and filtered to
      0.8-4.5Hz) from five events at the North Korean
      nuclear test site, showing strong inter-event
      correlation. By modeling this repeated structure, new
      events at this site can be detected even if only observed
      by a single station. }
    \label{fig:xcalign}
\end{figure}

\vspace{-.5em}
We model seismic events as point sources, localized in space and time,
that release energy in the form of seismic waves. These include
compression and shear (P and S) waves that
travel through the solid earth, as well as surface waves such as
Love and Rayleigh waves. Waves are further categorized
into {\em phases} according to the path traversed from the source
to a detecting station; for example, we distinguish P
waves propagating directly upwards to the surface (Pg phases)
from the same waves following a path guided along the crust-mantle
boundary (Pn phases), among other options. 

Given an event's location, it is possible to predict arrival times for
each phase by considering the length and characteristics of the event--station
path. Seismologists have developed a number of travel
time models, ranging from simple models that condition only on event
depth and event-station distance \citep{kennett1991traveltimes}, to more sophisticated models
that use an event's specific location to provide more
accurate predictions incorporating the local velocity structure
\citep{simmons2012llnl}. Inverting the predictions of a travel-time
model allows events to be located by triangulation given a set of arrival times. 

Seismic stations record continuous ground motion along one or more
axes,\footnote{This work considers vertical motion only.} including background noise from natural and human sources, as well as the
signals generated by arriving seismic phases. The detailed
fluctuations in these signals are a function of the source  as well as a path-dependent transfer function,
in which seismic energy is modulated and distorted by the geological
characteristics of the event--station path. Since geology does not
change much over time, events with similar
locations and depths tend to generate highly correlated waveforms (\Cref{fig:xcalign}); the lengthscale at which such
correlations are observed depends on the local geology and may range
from hundreds of meters up to tens of kilometers. 

\subsection{Seismic monitoring systems}

Traditional architectures for seismic monitoring operate via bottom-up
processing (\Cref{fig:monitoring_comparison}). The waveform at each station is thresholded to
produce a feature representation consisting of discrete ``detections'' of possible phase
arrivals. {\em
  Network processing} attempts to associate detections from
all stations into a coherent set of events. Potential complications include
false detections caused by noise, and missed detections from
arrivals for which the station processing failed to trigger.
In addition, the limited information (estimated arrival time, amplitude, and azimuth) in a single detection
means that an event typically requires detections from
at least three stations to be formed.

The NETVISA system \citep{arora2010global,arora2013net} replaces heuristic network
processing with Bayesian inference in a principled model of seismic
events and detections, including probabilities of false and missing
detections. Maximizing posterior probability in this model via
hill-climbing search yields an event bulletin that incorporates all available
data accounting for uncertainty. Because NETVISA separates 
inference from the construction of an explicit domain
model, domain experts can improve system
performance simply by refining the model.  Compared to
the IMS's previous network processing system (Global
Association, or GA), NETVISA provides significant improvements
in location accuracy and a 60\% reduction in missed events; as of this
writing, it has been proposed by the UN as the new production monitoring
system for the CTBT.

Recently, new monitoring approaches have been proposed using the
principle of {\em waveform matching}, which exploits correlations
between signals from nearby events to detect and
locate new events by matching incoming signals against a library of
historical signals. These promise the ability
to detect events up to an order of magnitude below the threshold of a
detection-based system \citep{gibbons2006detection,schaff2010one}, and
to locate such events even from a single station
\citep{schaff2012seismological}. However, adoption has been hampered by the inability to detect
events in locations with no historical seismicity, a crucial requirement
for nuclear monitoring. In addition, it has not been clear how to quantify
the reliability of events detected by waveform correlation, how to
reconcile correlation evidence from multiple stations,
or how to combine correlation and detection-based
methods in a principled way. Our work resolves these questions by
showing that both triangulation and waveform matching behaviors emerge
naturally during inference in a unified generative model of seismic
signals. 

\section{Modeling seismic waveforms}

\begin{figure}
\centering
    \includegraphics[width=0.45\textwidth]{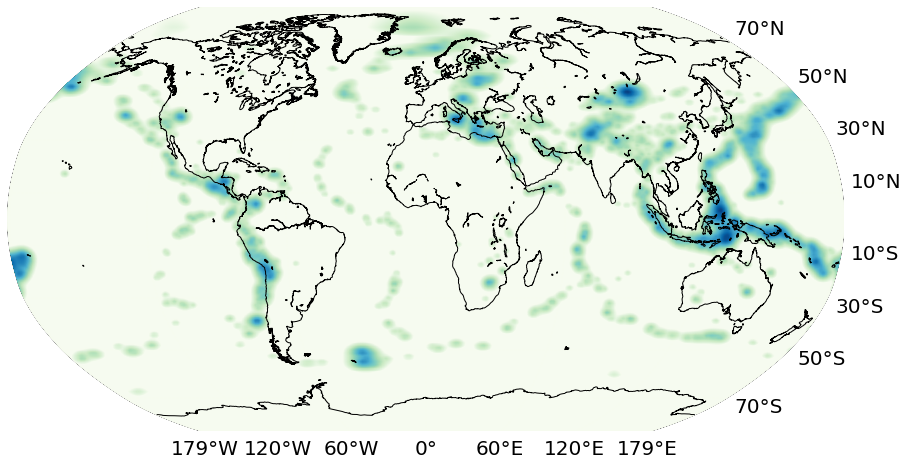}
    \caption{Global prior on seismic event locations.}
\label{fig:global_prior}
\end{figure}

\begin{figure}[tb]
\centering
 \begin{subfigure}[b]{.45\textwidth}
 \includegraphics[width=\textwidth]{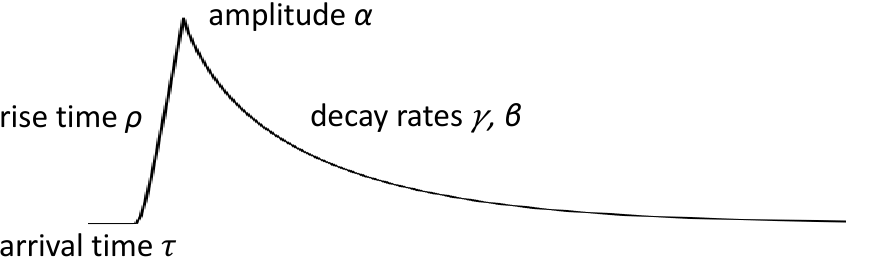}
 \caption{Parametric form $g(t; \v{\theta})$ modeling the signal
   envelope of a single arriving phase. (\cref{eqn:parametric_form})}
 \label{fig:parametric_form}
 \end{subfigure}

 \begin{subfigure}[b]{.45\textwidth}
 \includegraphics[width=\textwidth,height=3cm]{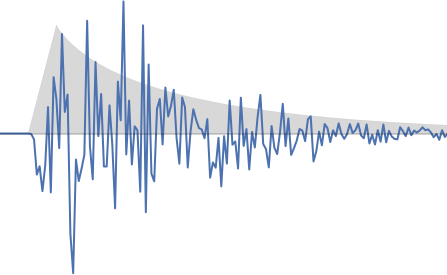}
 \caption{Signal for an arriving phase, generated by multiplying
   the parametric envelope $g$ (shaded) by a zero-mean modulation signal
   $m(t; \v{w})$. (\cref{eqn:modulation})}
 \label{fig:modulated_signal}
 \end{subfigure}

 \begin{subfigure}[b]{.45\textwidth}
 \includegraphics[width=\textwidth, height=3cm]{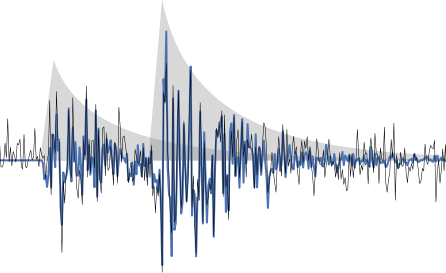}
 \caption{The final generated signal $\v{s}_j$ sums the contributions
   of all arriving phases with an autoregressive background
   noise process. (\cref{eqn:predicted_signal,eqn:final_signal})}
 \label{fig:overlapping_phases}
 \end{subfigure}
\caption{Steps of the SIGVISA forward model.}
\label{fig:model_algebra}
\end{figure}

SIGVISA (our current work) extends NETVISA by incorporating 
waveforms directly in the generative model, eliminating the need for bottom-up
detection processing. Our model describes a joint distribution
$p(\v{E}, \v{S}) = p(\v{S}|\v{E})p(\v{E})$ on $n_E$ seismic events $\v{E}$ and 
signals $\v{S}$ observed across $n_S$ stations. We model event occurrence as a time-homogenous Poisson process,
\begin{align*}
p(n_E) = \text{Poisson}(\lambda T),\; p(\v{E}) = p(n_E) n_E! \prod_{i=1}^{n_E} p(\v{e}_i),
\end{align*}
so that the number of events generated during a time period of
length $T$ is itself random, and each event is sampled independently from
a prior $p(\v{e}_i)$ over surface location, depth, origin time, and
magnitude. The homogeneous process implies a uniform prior on origin
times, with a labeling symmetry that we correct by multiplying by the
permutation count $n_E!$. Location and depth priors, along with the event
rate $\lambda$, are estimated from historical seismicity as described by
\citet{arora2013net}. The location prior is a mixture of a kernel density
estimate of historical events, and a uniform component to allow
explosions and other events in locations with no previous seismicity (\Cref{fig:global_prior}).

We assume that signals at different stations are
conditionally independent, given events, and introduce auxiliary
variables $\v{\theta}$ and $\v{w}$ governing signal generation at each
station, so that the forward model has the form
\begin{equation}
p(\v{S}|\v{E}) = \prod_{j=1}^{n_S} \left(\iint p(\v{s}_j |
  \v{\theta}_j, \v{w}_j)p(\v{\theta}_j|\v{E}) p(\v{w}_j| \v{E})
  d\v{w}_j d\v{\theta}_j \right).
\end{equation}
The parameters $\v{\theta}$ describe an
envelope shape for each arriving phase, and $\v{w}$ describes {\em
  repeatable} modulation processes
that multiply these envelopes to generate observed waveforms (\Cref{fig:model_algebra}). In
particular, we decompose these into independent (conditioned on event
locations) components $\v{\theta}_{i,j,k}$ and $\v{w}_{i,j,k}$
describing the arrival of phase $k$ of event $i$ at station $j$. 

The envelope of each phase is modeled by a linear onset
followed by a poly-exponential decay (Figure~\ref{fig:parametric_form}), 
\begin{equation}g(t; \theta_{i,j,k}) = \left\{\begin{array}{ll}
0 & \text{ if } t \le \tau\\
\alpha (t-\tau) / \rho & \text{ if } \tau < t \le \tau+\rho\\
\alpha (t-\tau+1)^{-\gamma} e^{-\beta (t-\tau)} &\text{ otherwise}\\
\end{array}\right.\label{eqn:parametric_form}
\end{equation}
with parameters $\theta_{i,j,k} = (\tau, \rho, \alpha, \gamma,
\beta)_{i,j,k}$ consisting of an arrival time $\tau$, rise time $\rho$, amplitude
$\alpha$, and decay rates $\gamma$ and $\beta$ governing respectively
the envelope peak and its coda, or long-run decay. This decay form is inspired by previous work modeling seismic
coda \citep{mayeda2003stable}, while the linear
onset follows that used by 
\citet{cua2005creating} for seismic early warning. 

\begin{figure}
\centering
\begin{subfigure}[b]{0.235\textwidth}
    \includegraphics[width=\textwidth]{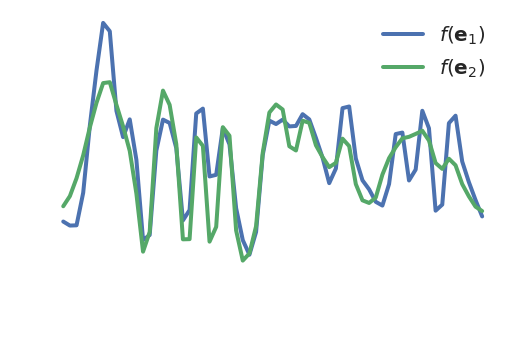}
  \caption{Co-located sources.}
\end{subfigure}
\begin{subfigure}[b]{0.235\textwidth}
    \includegraphics[width=\textwidth]{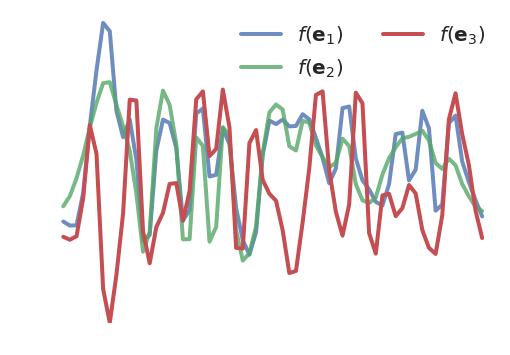}
  \caption{Adding a distant source.}
\end{subfigure}
\caption{Samples from a GP prior on db4 wavelets.}
\label{fig:wavelet_gps}
\end{figure}

To produce a signal, the envelope is multiplied by a {\em modulation
  process} $m$ (Figure~\ref{fig:modulated_signal}), parameterized
by wavelet coefficients $\v{w}_{i,j,k}$ so that
\begin{equation}
m(t; \v{w}_{i,j,k}) = \left\{\begin{array}{ll} (\mathbf{D}\v{w}_{i,j,k})(t) & \text{if } 0 \le t <
                                                    20s\\\ve(t) &
                                                                  \text{otherwise}\end{array}\right.
\label{eqn:modulation}
\end{equation}
where $\mathbf{D}$ is a discrete wavelet transform matrix, and
$\ve(t)\sim\N(0, 1)$ is a Gaussian white noise process. We
explicitly represent coefficients describing the first 20
seconds\footnote{This cutoff was chosen to
  capture repeatability of the initial arrival period, which is
  typically the most clearly observed, while still fitting historical
  models in memory.} of each 
arrival, and model the modulation as random after that point. We use an order-4 Daubechies wavelet
basis \citep{daubechies1992ten}, so that for 10Hz signals each
$\v{w}_{i,j,k}$ is a vector of 220 coefficients.  As described below, we
model $\v{w}$ jointly across events using a Gaussian process, so that
our modulation processes are {\em repeatable}: events in nearby locations
will generate correlated signals (\Cref{fig:wavelet_gps}). 

Summing the signals from all arriving phases yields the {\em
  predicted signal} $\v{\bar{s}}_j$,
\begin{equation}
\bar{s}_j(t) = \sum_{i, k} g(t; \v{\theta}_{i,j,k})\cdot
m(t-\tau_{i,j,k}; \v{w}_{i,j,k});
\label{eqn:predicted_signal}
\end{equation}
and we generate the observed signal $\v{s_j}$ (\Cref{fig:overlapping_phases})
by adding an order-$R$ autoregressive noise process,
\begin{align}
p(\v{s}_j | \v{\theta}_j, \v{w}_j) &= p_{AR}(\v{s}_j - \v{\bar{s}}_j - \mu_j;
\sigma^2_j, \v{\phi}_j)\label{eqn:final_signal}\\
p_{AR}(\v{z}; \sigma^2, \v{\phi}) &= \prod_{t=1}^T \N\left(z(t);
\sum_{r=1}^R \v{\phi}_r z(t-r), \sigma^2\right).\nonumber
\end{align}
We adapt the noise process online during inference, following station-specific priors on the mean $p(\mu_j)$, variance
$p(\sigma^2_j)$, and autoregressive coefficients $p(\v{\phi}_j)$.

\subsection{Repeatable signal descriptions}



For each station $j$ and phase $k$, we model the signal
descriptions $p(\v{\theta}_{j,k} | \v{E})$ and $p(\v{w}_{j,k} |
\v{E})$ jointly across events as Gaussian process 
transformations of the event space $\v{E}$ \citep{rasmussen2006}, so that events in
nearby locations will tend to generate similar envelope shapes and
correlated modulation signals, allowing our system to detect and
locate repeated events even from a weak signal recorded at a single
station. Since the events are unobserved, this is effectively a
GP latent variable model \citep{lawrence2004gaussian}, with the twist that in our model the
outputs $\v{\theta}, \v{w}$ are themselves latent variables, observed
only indirectly through the signal model $p(\v{S} | \v{\theta}, \v{w})$. 


We borrow models from geophysics to predict the arrival time and
amplitude of each phase: given the origin time, depth, and
event-station distance, the IASPEI-91 \citep{kennett1991traveltimes} travel-time
model predicts an arrival time $\bar{\tau}$, and the Brune
source model \citep{brune1970tectonic} predicts a source (log) amplitude
$\log \bar{\alpha}$ as a function of event magnitude. We then use
GPs to model deviations from these physics-based predictions; that is,
we take $\bar{\tau}$ and $\log \bar{\alpha}$ as the
mean functions for GPs modeling $\tau$ and $\log
\alpha$ respectively. The remaining shape parameters $\rho, \gamma,
\beta$ (in log space) and the 220 wavelet coefficients in $\v{w}_{i,j,k}$ are 
each modeled by independent zero-mean GPs. 

\begin{table}
\centering
\begin{tabular}{ll}
\hline
\textbf{Parameter} & \textbf{Features $\phi(\v{e})$} \\\hline
Arrival time ($\tau$) & n/a  \\
Amplitude (log $\alpha$) & $\left(1,\Delta,\sin(\frac{\Delta}{15000}),
                           \cos(\frac{\Delta}{15000})\right)$ \\
Onset (log $\rho$) & (1, mb) \\
Peak decay (log $\gamma$) & (1, mb, $\Delta$) \\
Coda decay (log $\beta$) & (1, mb, $\Delta$) \\
Wavelet coefs ($w$) & n/a \\\hline
\end{tabular}
\caption{Feature representations in terms
  of magnitude mb and event--station distance $\Delta$ (km).}
\label{tbl:gp_models}
\end{table}

All of our GPs share a common covariance form,
\[k(\v{e}, \v{e}') = \phi(\v{e})^T \v{B}\phi(\v{e}') + \sigma^2_f
k_\text{Mat\'ern}(d_\ell(\v{e}, \v{e}')) + \sigma^2_n \delta,\]
where the first term models an unknown function that is linear in some
feature representation $\phi$, chosen separately for each parameter
(\Cref{tbl:gp_models}). This represents general regularities such as
distance decay that we expect to hold even in regions with no
observed training data. For efficiency and to avoid degenerate
covariances we represent this component in weight space \citep[section 2.7]{rasmussen2006}; at test time we choose $\v{B}$
to be the posterior covariance given training
data. The second term models an unknown
differentiable function using a stationary Mat\'ern ($\nu=3/2$) kernel \citep[Chapter
4]{rasmussen2006}, with great-circle distance metric controlled by
lengthscale hyperparameters $\ell$; this allows our model to represent
detailed local seismic structure. We also include iid noise
$\sigma^2_n \delta$ to encode unmodeled variation
between events in the same location.

To enable efficient training and test-time GP predictions,
we partition the training data into spatially local regions using
$k$-means clustering, and factor the nonparametric (Mat\'ern)
component into independent regional models. This allows predictions in
each region to be made efficiently using only a small number of nearby events, avoiding a na\"ive
$O(n^2)$ dependence on the entire training set. Each region is given
separate hyperparameters, allowing our models to adapt to spatially varying seismicity.

\subsection{Collapsed signal model}

The explicit model described thus far exhibits tight coupling between
envelope parameters and modulation coefficients; a small shift in a
phase's arrival time may significantly change the wavelets needed to
explain an observed signal, causing inference moves that do not
account for this joint structure to fail. Fortunately, it is possible to exactly
marginalize out the coefficients $\v{w}$ so that they do not need to
be represented in inference. This follows from the linear Gaussian
structure of our signal model: GPs induce a Gaussian distribution on
wavelet coefficients, which are observed under linear projection (a
wavelet transform followed by envelope scaling) with autoregressive
background noise. Thus, the collapsed distribution
$p(\v{s}_j | \v{\theta}_j, \v{E}) = \int p(\v{s}_j | \v{\theta}_j, \v{w}_j)
p(\v{w}_j | \v{E}) d\v{w}_j$
is multivariate Gaussian, and in principle can be evaluated directly. 

Doing this efficiently in practice requires exploiting graphical model
structure. Specifically, we formulate the signal model at each station 
as a linear Gaussian state space model, with a state vector
that tracks the AR noise process as well as the set of
wavelet coefficients that actively contribute to the
signal. Due to the recursive structure of wavelet bases, the number
of such coefficients is only logarithmic at each timestep. We exploit
this structure, the same used
by fast wavelet transforms, to efficiently\footnote{Requiring time linear in the signal length:
  $O(T (K \log C + R)^2 )$ for a signal of length $T$ with order-$R$
  AR noise and at most $K$ simultaneous phase arrivals, each described
  by $C$ wavelet coefficients.} compute coefficient posteriors and
marginal likelihoods by Kalman filtering
\citep{grewal2014kalman}. 

We assume for efficiency that test-time signals from different
events are independent given the training data. During training, however, it is necessary to compute
the joint density of signals involving multiple events so that we can
find correct alignments. This requires us to
pass messages $f_j(\v{w}_j) =
p(\v{s}_j | \v{w}_j, \v{\theta}_j)$ from each signal upwards to the GP
prior  \citep{koller2009probabilistic}. We compute a diagonal
approximation \[\tilde{f}_j(\v{w}_j) = \frac{1}{Z_j}\prod_{c=1}^{220} \N(\v{w}_{j,c};
\tilde{\nu}_{j,c}, \tilde{\xi}_{j,c})\] by dividing the Kalman
filtering posterior on wavelet coefficients by the (diagonal Gaussian)
prior. The product of these messages with the GP priors $p(\v{w}_c | \v{E})
\sim \N(\v{\mu}_c(\v{E}), \v{K}_c(\v{E}) )$, integrated over coefficients $\v{w}$, gives an approximate
joint density
\begin{equation}
p(\v{S} | \v{\theta}, \v{E}) \approx \left(\prod_{j=1}^N
  \frac{1}{Z_j} \right) \prod_c \N\left(\bar{\nu}_c; \v{\mu}_c(\v{E}),   \v{K}_c(\v{E}) + \v{\tilde{\xi}}_{c} \right)\label{eqn:efficient_joint_density},
\end{equation}
in which the wavelet GPs are evaluated at the values of (and
with added variance given by) the approximate messages from observed signals. We target this
approximate joint density during training, and condition our test-time
models on the upwards messages generated by training signals. 

\section{Training}

We train using a bulletin of historical event locations which
we take as ground truth (relaxing this assumption to incorporate noisy training
  locations, or even fully unsupervised training, is important
  future work). Given observed events, we use
training signals to estimate the GP models---hyperparameters as well as the
  upwards messages from training signals---along with priors
  $p(\mu_j)$, $p(\sigma^2_j)$, and $p(\v{\phi}_j)$ on background
  noise processes at each station. We use the EM algorithm
  \citep{dempster1977maximum} to search for maximum likelihood parameters integrating over
  the unobserved noise processes and envelope shapes. 

  The E step runs MCMC inference (\Cref{sec:inference}) to sample from
  the posterior over the latent variables
  $\v{\theta}, \v{\mu}, \v{\sigma}^2, \v{\phi}$ under the collapsed
  objective described above. We approximate the sampled
  posterior on $\v{\theta}$ by univariate Gaussians to compute
  approximate upwards messages. These are used in the M step to fit GP hyperparameters via
  gradient-based optimization of the marginal likelihood
  (\ref{eqn:efficient_joint_density}). The priors on background noise means
  $p(\mu_j)$ and coefficients $p(\v{\phi}_j)$ are fit as Gaussian and
  multivariate Gaussian respectively. For the noise variance
  $\sigma^2_j$ at each station, we fit log-normal, inverse Gamma, and
  truncated Gaussian priors and select the most likely; this adapts
  for different noise distributions between stations.


\section{Inference}
\label{sec:inference}
We perform inference using reversible jump MCMC \citep{hastie2012model} applied
to the collapsed model. Our algorithm consists of a cyclic sweep of
single-site, random-walk Metropolis-Hastings moves over all currently
instantiated envelope parameters $\v{\theta}$, autoregressive noise
parameters $\mu, \sigma^2, \phi$ at each station, and event
descriptions $(\v{e}_i)_{i=1}^{n_E}$ including surface location,
depth, time, and magnitude. We also include custom
moves that propose swapping the associations of consecutive arrivals,
aligning observed signals with GP predicted signals,
and shifting envelope peak times to match those of the observed signal. 

To improve event mixing, we augment our model to include {\em unassociated
  arrivals}: phase arrivals not generated by any particular event,
with envelope parameters and modulation from a fixed Gaussian
prior. Unassociated arrivals are useful in allowing events to be built
and destroyed piecewise, so that we are not required to perfectly
propose an event and all of its phases in a single shot. They can also be
viewed as small events whose locations have been integrated out. Our birth
proposal generates unassociated arrivals with probability proportional
to the signal envelope, so that periods of high signal energy are
quickly explained by unassociated arrivals which may then be
associated into larger events.\footnote{In this sense the unassociated
arrivals play a role similar to detections produced by traditional
station processing. However, they are not generated by bottom-up
preprocessing, but as part of a dynamic inference procedure that may
create and destroy them using information from observed signals as well as top-down
event hypotheses.}

Event birth moves are constructed using two complementary proposals. The first is based on a
Hough transform of unassociated arrivals; it grids the 5D event space
(longitude, latitude, depth, time, magnitude) and scores each bin
using the log likelihood of arrivals greedily associated with an event
in that bin. The second proposal is a mixture of Gaussians centered at
the training events, with weights determined by waveform
correlations against test signals. This allows us to recover weak events
that correlate with training signals, while the Hough proposal can construct events in regions with no previous seismicity. 

In proposing a new event we must also propose envelope parameters for
all of its phases. Each phase may associate a currently unassociated arrival; where
there are no plausible arrivals to associate we parent-sample envelope
parameters given the proposed event, then run auxiliary Metropolis-Hastings steps
\citep{storvik2011} to adapt the envelopes to observed signals. The
proposed event, associations, and envelope parameters are jointly accepted or
rejected by a final MH step. Event death moves similarly involve
jointly proposing an event to kill along with a set of phase arrivals
to delete (with the remainder preserved as unassociated). 

By chaining birth and death moves we also construct
mode-jumping moves that repropose an existing event, along with split
and merge moves that replace two existing events with a single one, and
vice versa. \citet{moore2016signal} describes our inference moves in
more detail.

\section{Evaluation}
\vspace{-0.5em}

\begin{figure}
\centering
\includegraphics[width=.45\textwidth]{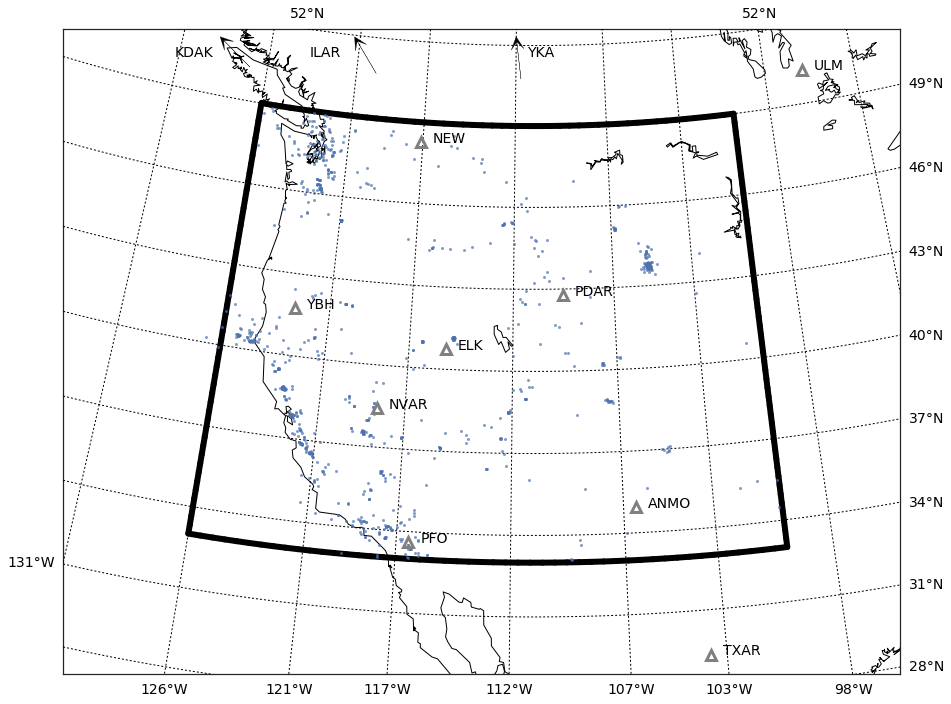}
\caption{Training events (blue dots) from the western US dataset, with region of interest outlined. Triangles indicate IMS stations.}
\label{fig:iscevents}
\end{figure}

\begin{figure}
\centering
    \includegraphics[width=0.5\textwidth]{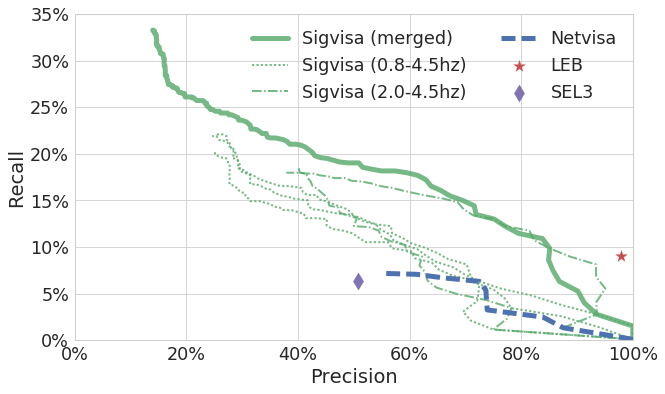}
    \caption{Precision-recall performance over the two-week test
      period, relative to the reference bulletin.}
  \label{fig:test_prec_recall}
\end{figure}

\begin{figure}
\centering
    \includegraphics[width=0.45\textwidth]{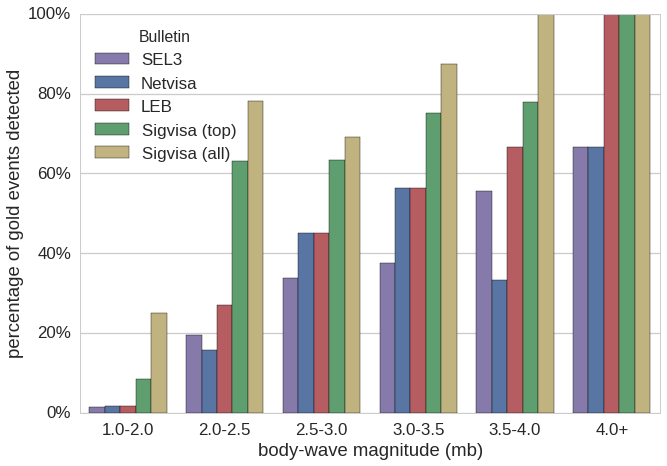}
    \caption{Event recall by magnitude range. The SIGVISA (top) bulletin is defined to match the
      precision of SEL3 (51\%).}
  \label{fig:isc_evs_by_mb}
\end{figure}

\begin{figure*}
\centering
\begin{subfigure}[b]{0.30\textwidth}
\stackinset{r}{}{b}{}{\includegraphics[width=0.75in]{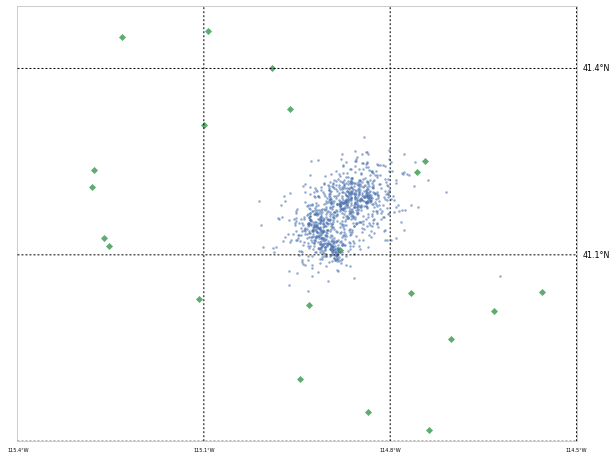}}
  {\includegraphics[width=\textwidth]{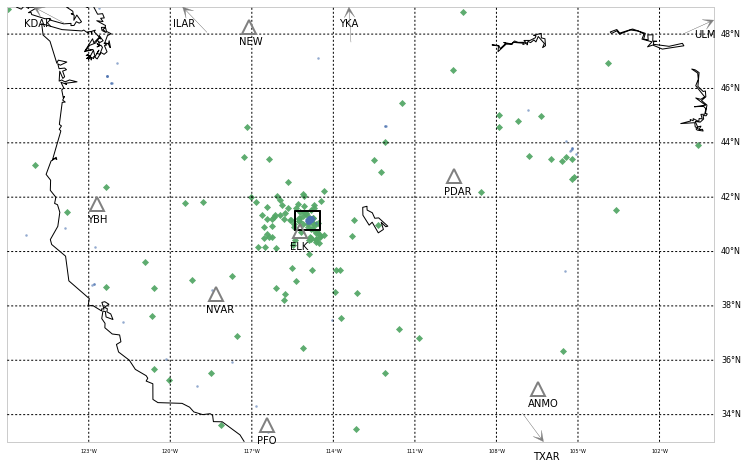}}
\caption{NETVISA (139 events)}
\label{fig:visa_map}
\end{subfigure}
\begin{subfigure}[b]{0.30\textwidth}
\stackinset{r}{}{b}{}{\includegraphics[width=0.75in]{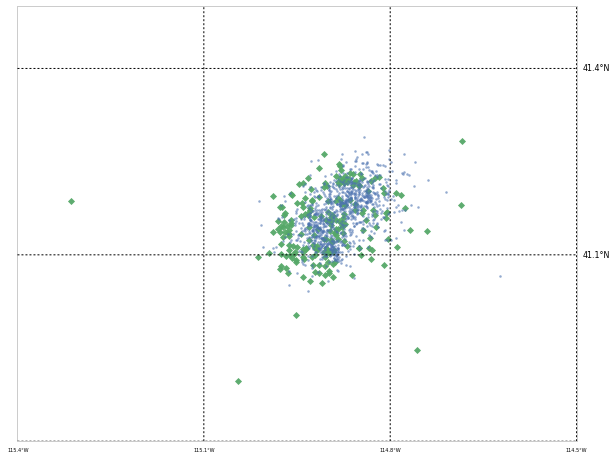}}
  {\includegraphics[width=\textwidth]{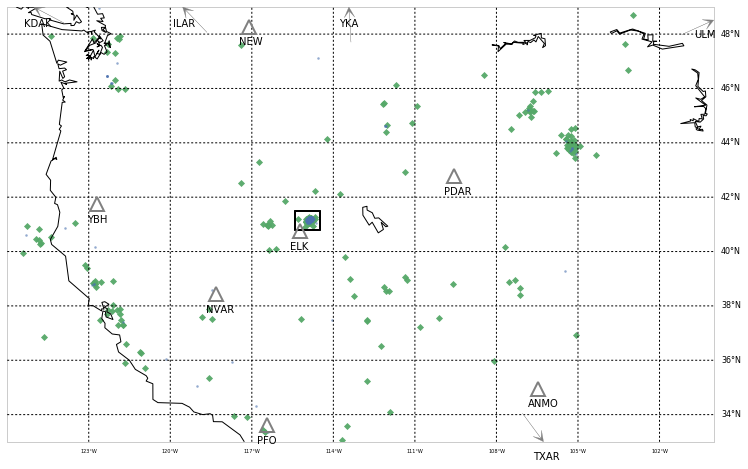}}
\caption{SIGVISA top events (393 events)}
\label{fig:sigvisa_map}
\end{subfigure}
\begin{subfigure}[b]{0.30\textwidth}
    \includegraphics[width=\textwidth]{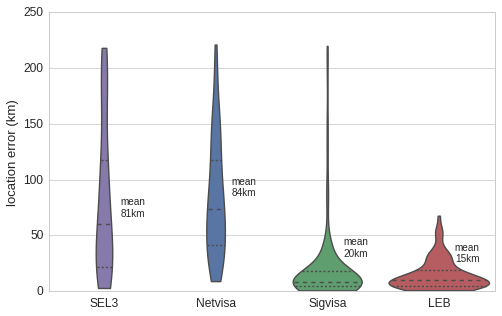}
    \caption{Distribution of location errors.}
  \label{fig:test_location_err}
\end{subfigure}
\caption{Inferred events (green), with inset close-up of Wells
  aftershocks. Reference events are in blue.}
\label{fig:inferred_map}
\end{figure*}

We consider the task of monitoring seismic events in the western
United States, which contains both significant natural seismicity and
regular mining explosions. We focus in particular on the time period
immediately following the magnitude 6.0 earthquake near Wells, NV, on
February 21, 2008, which generated a large number of aftershocks. We
train on one year of historical data  (\Cref{fig:iscevents}), from January to December 2008; to enable SIGVISA to recognize aftershocks using waveform
correlation, we also train on the first six hours following the Wells mainshock. The test period is two weeks long, beginning twelve hours
after the Wells mainshock; the six hours immediately preceding were used as a validation set.

We compare SIGVISA's performance to that of existing systems that also
process data from the International Monitoring System.  SEL3 is the
final-stage automated bulletin from the CTBTO's existing system (GA); it is reviewed by a team of human analysts to
produce the Late Event Bulletin (LEB) reported to the member
states. We also compare to the automated NETVISA bulletin\citep{arora2013net}, which
implements detection-based Bayesian monitoring. 

We construct a reference bulletin by combining events from regional networks aggregated by the
National Earthquake Information Center (NEIC),
and an analysis of aftershocks from the Wells earthquake based on data
from the transportable US Array and 
temporary instruments deployed by the University of Nevada, Reno (UNR)
\citep{smith2011preliminary}. Because the reference bulletin has access to many 
sensors not included in the IMS, it is a plausible source of ``ground
truth'' to evaluate the IMS-based systems. 

We trained two sets of SIGVISA models, on
broadband (0.8-4.5Hz) signals as well as a higher-frequency band
(2.0-4.5Hz) intended to provide clearer evidence of regional
events, using events observed from the reference bulletin. 
To produce a test bulletin, we ran three MCMC chains on
broadband signals and two on high-frequency signals, and merged the
results using a greedy heuristic that iteratively selects the
highest-scoring event from any chain excluding duplicates. Each
individual chain was parallelized by dividing the test period into 168
two-hour blocks and running independent inference on each block of
signals. Overall SIGVISA inference used 840 cores for 48 hours.

We evaluate each system by computing a minimum weight maximum
cardinality matching between the inferred and reference bulletins, in a bipartite
graph with edges weighted by distance and restricted to
events separated by at most $2\deg$ in distance and 50s in
time. Using this matching, we report precision (the percentage of inferred
events that are real), recall (the percentage of real events detected
by each system), and mean location error of matched events. For
NETVISA and SIGVISA, which attach a confidence score to each event, we
report a precision-recall curve parameterized by the confidence
threshold (\Cref{fig:test_prec_recall}). 

Our results show that the merged SIGVISA
bulletin dominates both NETVISA and SEL3. When operating at the same
precision as SEL3 (51\%), SIGVISA achieves recall three times higher
than SEL3  (19.3\% vs 6.4\%), also eclipsing the 7.3\% recall achieved by NETVISA at a slightly higher
precision (54.7\%). The human-reviewed LEB achieves near-perfect precision
but only 9\% recall, confirming that many events recovered by SIGVISA
are not obvious to human analysts. The most sensitive SIGVISA bulletin
recovers a full 33\% of the reference events, at the cost of
many more false events (14\% precision). 

Signal-based modeling particularly improves recall for low-magnitude
events (\Cref{fig:isc_evs_by_mb}) for which
bottom-up processing may not register detections. We also observe
improved locations (\Cref{fig:inferred_map})
for clusters such as the Wells aftershock sequence where observed
waveforms can be matched against training data. 

\begin{figure}[tb]
\centering
\begin{subfigure}[b]{0.45\textwidth}
    \includegraphics[width=\textwidth]{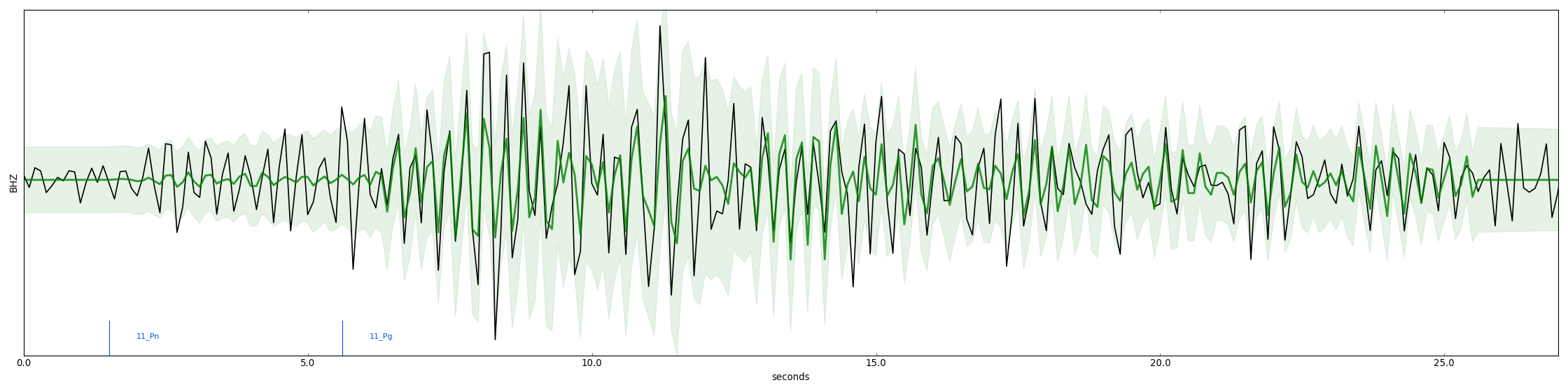}
    \caption{Likely mining explosion at Black Thunder Mine. Location 105.21$\deg$ W, 43.75$\deg$ N, depth 1.9km, origin time 17:15:58 UTC, 2008-02-27, mb 2.6, recorded at PDAR (PD31).}
\end{subfigure}

\begin{subfigure}[b]{0.45\textwidth}
    \includegraphics[width=\textwidth]{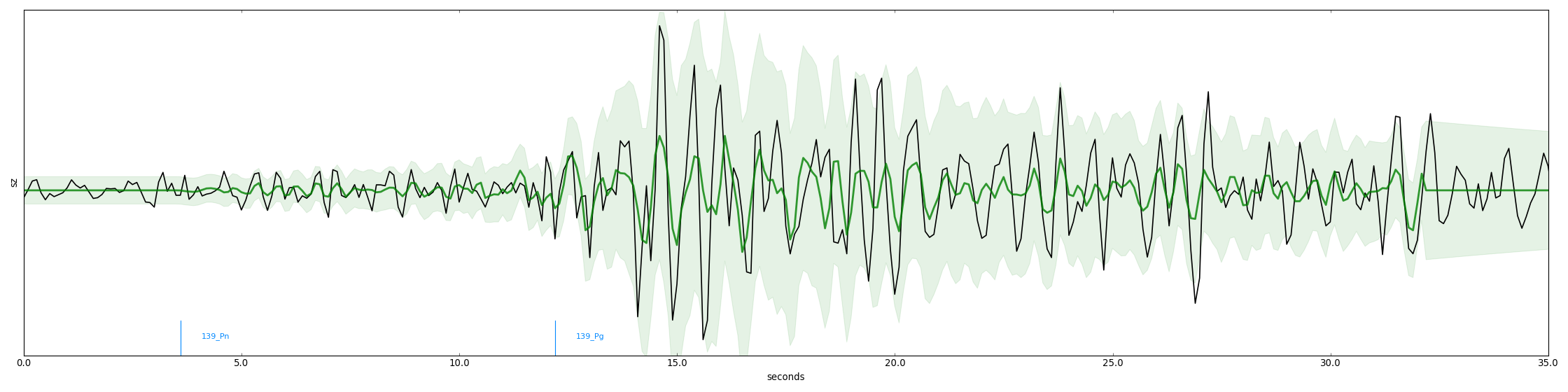}
    \caption{Event near Cloverdale, CA along the Rodgers Creek
      fault. Location 122.79$\deg$ W, 38.80$\deg$ N, depth 1.6km,
      origin time 05:20:56 UTC, 2008-02-29,  mb 2.6, recorded at NVAR
      (NV01).}
\end{subfigure}
\caption{Waveform evidence for two events detected by SIGVISA
  but not the reference bulletin. Green indicates the model
  predicted signal (shaded $\pm 2\sigma$); black is the observed
signal (filtered 0.8-4.5Hz).}
  \label{fig:sigvisa_genuine_evs}
\end{figure}

Interestingly, because we do not have access to absolute ground
truth, some events labeled as false in our evaluation may actually be genuine events missed
by the reference bulletin. \Cref{fig:sigvisa_genuine_evs} shows two
candidates for such events, with strong correspondence between the model-predicted and observed
waveforms. The existence of such events provides reason to believe
that SIGVISA's true performance on this dataset is modestly higher than our evaluation
suggests.

\subsection{{\em de novo} events}

\begin{figure}
\centering
\includegraphics[width=0.45\textwidth]{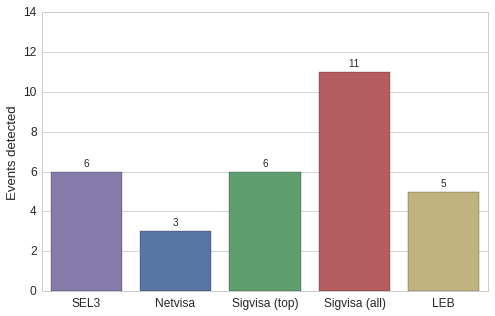}
\caption{Recall for 24 {\em de novo} events between January and March
2008. }
\label{fig:denovo_results}
\end{figure}

For nuclear monitoring it is particularly important
to detect {\em de novo} events: those with no nearby historical
seismicity. We define ``nearby'' as within 50km. Our two-week test period includes only three
such events, so we broaden the scope to the
three-month period of January through March 31, 2008, which
includes 24 de novo events. We evaluated each system's recall
specifically on this set: of the 24 {\em de novo} reference events, how many were detected? 

As shown in \Cref{fig:denovo_results}, SIGVISA's
performance matches or exceeds the other systems. Operating at the
same precision as SEL3, it detects the same number (6/24) of de novo 
events. This suggests that SIGVISA's
improved performance on repeated events---including almost
all of the natural seismicity in the western US during our two-week
test period---does not come at a cost for de novo events. To the contrary,
the full high-sensitivity SIGVISA bulletin includes six genuine events
missed by all other IMS-based systems.

\vspace{-1em}
\section{Discussion}
\vspace{-1em}
Our results demonstrate the promise of 
Bayesian inference on raw waveforms for monitoring seismic
events. Applying MCMC inference to a generative probability model of repeatable
seismic signals, we recover up to three times as many events as a
detection-based baseline (SEL3) while operating at the same precision,
and reduce mean location errors by a factor of four while greatly increasing
sensitivity to low-magnitude events. Our system maintains effective
performance even for events in regions with no historical seismicity
in the training set.

A major advantage of the generative formulation is that the explicit
model is interpretable by domain experts. We continue to engage with
seismologists on potential model improvements, including tomographic
travel-time models, directional information from seismic arrays
and horizontal ground motion, and explicit modeling of earthquake
versus explosion sources. Additional directions include
more precise investigations of our model's ability to quantify
uncertainty and to estimate its own detection limits as a function of
network coverage and historical seismicity. We also expect to continue scaling to global seismic data,
exploiting parallelism and refining our inference moves and
implementation. More generally, we hope that successful application of complex 
Bayesian models will inspire advances in probabilistic
programming systems to make generative modeling accessible to a wider
scientific audience.

\vspace{-0.5em}
\subsubsection*{Acknowledgements}
\vspace{-0.5em}
We are very grateful to Steve Myers (LLNL) and Kevin Mayeda
(Berkeley/AFTAC) for sharing their expertise on seismic modeling and
monitoring and advising on the experimental setup. This work is
supported by the Defense Threat Research Agency (DTRA) under grant
\#HDTRA-1111-0026, and experiments by an Azure for Research grant from Microsoft.

\newpage
\bibliographystyle{apa}
\bibliography{references}

\end{document}